\documentclass[runningheads]{llncs}
\usepackage{hyperref}
\hypersetup{
    colorlinks=true,
    linkcolor=blue,
    filecolor=magenta,      
    urlcolor=cyan
    }
\usepackage{graphicx}
\usepackage{booktabs}
\usepackage{rotating}
\usepackage{multirow}
\usepackage{amsmath}
\usepackage{xcolor}
\usepackage{amssymb}
\definecolor{red}{RGB}{255,0,0}
\definecolor{brown}{RGB}{139,69,19}
\definecolor{green}{RGB}{0,128,0}
\definecolor{blue}{RGB}{0,0,255}

\newcommand{\blue}[1]{{\color{blue} #1}}
\newcommand{\brown}[1]{{\color{brown} #1}}
\newcommand{\sstitle}[1]{\vspace{1mm} \noindent {\bf #1}}
\newcommand{\dashed}{--}
\begin{document}
\title{Do Similar Entities have Similar Embeddings?}

%
%\titlerunning{Abbreviated paper title}
% If the paper title is too long for the running head, you can set
% an abbreviated paper title here
%
\author{
Nicolas Hubert\inst{1,2}\orcidID{0000-0002-4682-422X} \and
Heiko Paulheim\inst{3}\orcidID{0000-0003-4386-8195} \and
Armelle Brun\inst{2}\orcidID{0000-0002-9876-6906} 
\and
Davy Monticolo\inst{1}\orcidID{0000-0002-4244-684X}
}

\authorrunning{N. Hubert et al.}

\institute{Université de Lorraine, ERPI, Nancy, France \and
Université de Lorraine, CNRS, LORIA, Nancy, France \and
Data and Web Science Group, University of Mannheim, Germany
\email{\{nicolas.hubert,armelle.brun,davy.monticolo\}@univ-lorraine.fr}
\email{heiko@informatik.uni-mannheim.de}}
\maketitle              % typeset the header of the contribution
\begin{abstract}
Knowledge graph embedding models (KGEMs) developed for link prediction learn vector representations for entities in a knowledge graph, known as embeddings. A common tacit assumption is the \emph{KGE entity similarity assumption}, which states that these KGEMs retain the graph's structure within their embedding space, \textit{i.e.}, position similar entities within the graph close to one another. This desirable property make KGEMs widely used in downstream tasks such as recommender systems or drug repurposing. Yet, the relation of entity similarity and similarity in the embedding space has rarely been formally evaluated. Typically, KGEMs are assessed based on their sole link prediction capabilities, using ranked-based metrics such as Hits@K or Mean Rank. This paper challenges the prevailing assumption that entity similarity in the graph is inherently mirrored in the embedding space. Therefore, we conduct extensive experiments to measure the capability of KGEMs to cluster similar entities together, and investigate the nature of the underlying factors. Moreover, we study if different KGEMs expose a different notion of similarity. Datasets, pre-trained embeddings and code are available at: \url{https://github.com/nicolas-hbt/similar-embeddings/}.

\end{abstract}

\section{Introduction}
Knowledge Graphs (KGs) such as DBpedia~\cite{dbpedia} and YAGO~\cite{yago} represent facts as triples $(s,p,o)$ consisting of a subject $s$ and an object $o$ connected by a predicate $p$ defining their relationship.
Common learning tasks with KGs include entity clustering, node classification, and link prediction.

These tasks are predominantly tackled using Knowledge Graph Embedding Models (KGEMs), which generate dense vector representations for entities and relations of a KG, a.k.a. Knowledge Graph Embeddings (KGEs). The dense numerical vectors that are learnt for entities and relations are expected to preserve the intrinsic semantics of the KG~\cite{wang2021survey}. 

As KGEMs take into account the semantic relationship between two entities to learn embeddings, it is often taken for granted that the resulting embeddings capture both the semantics and attributes of entities and their relationships in the KG. Embeddings are thus widely used to measure the semantic similarity between entities and relations, facilitating data integration through entity or relation alignments~\cite{chen2019similarity,kalo2019similarity}. They are also used in various similarity-based tasks including entity similarity~\cite{sun2020entity} and conceptual clustering~\cite{gadelrab2020conceptual}.

However, the widespread assumption that KGEMs create semantically meaningful representations of the underlying entities (\textit{i.e.}, project similar entities closer than dissimilar ones) has been challenged recently~\cite{jain2021embeddings}. Jain \textit{et al.}~\cite{jain2021embeddings} demonstrate that entity embeddings learnt with KGEMs are not well-suited to identify the concepts or classes for a vast majority of KG entities, while simple statistical approaches provide comparable or better performance. Concerned with word embeddings, Ilievski \textit{et al.}~\cite{ilievski2023swj} also point out that KGEMs are consistently outperformed by simpler heuristics for similarity-based tasks. The authors argue that many properties on which KGEMs heavily rely on are not useful for determining similarity, which introduces noise and subsequently decreases performance. Our work falls within this line of thought. More specifically, we formulate the following research question:
\vspace{-0.025cm}
\begin{description}
    \item[RQ1.] To what extent does proximity in the embedding space align with the notion of entity similarity in the KG?
\end{description}
\vspace{-0.025cm}
We call this the \emph{KGE entity similarity assumption}. Notably, there is no universally accepted definition for entity similarity. In this work, we follow a straightforward approach: two entities in a KG are similar if we make similar statements about them. This aligns with the assumption of distributional semantics: words appearing in similar contexts are semantically similar. Answering \textbf{RQ1} requires remembering that most embedding-based models are trained to maximize rank-based metrics for link prediction, which disregards semantics. One could then argue that maximizing such metrics is at least partially decoupled from the task of learning similar vectors for similar entities. However, as related entities are more likely to appear in similar triples (\textit{e.g.}, featuring the same predicate), it is reasonable to believe that semantically close entities -- especially those connected to other entities through a shared set of predicates -- are also more likely to be assigned similar vectors~\cite{portisch2022knowledge}. This raises our second research question:
\vspace{-0.025cm}
\begin{description}
    \item[RQ2.] How do traditional rank-based metrics correlate with entity similarity?
\end{description}
\vspace{-0.025cm}
In other words, we ask whether KGEMs with good link prediction performance w.r.t. metrics such as Mean Reciprocal Rank (MRR) and Hits@$K$ necessarily group similar entities close in the embedding space. If so, it would be sensible to study which of the two is most likely to influence the other, and whether a form of causality (rather than just correlation) between these two aspects exists.

To delve further, it is worth noting that many link prediction train sets have been shown to suffer from extremely skewed distributions, especially in the occurrence of a subset of predicates~\cite{rossi2020relations}.
Rossi~\textit{et al.}~\cite{rossi2020relations} demonstrate that relying on global metrics (\textit{e.g.} Hits@$K$ and MRR) over such heavily skewed distributions hinders our understanding of KGEMs. Consequently, we ultimately distance ourselves from analyzing rank-based metrics, and dive deeper into entity embeddings with the sole consideration of studying how and why they may differ between models. In line with Rossi~\textit{et al.}~\cite{rossi2020relations} findings that a given subset of predicates is often likely to heavily influence entity representations -- we formulate our last research question as follows:
\vspace{-0.025cm}
\begin{description}
    \item[RQ3.] Do different KGEMs focus on different predicates to capture the notion of similarity in the embedding space?
\end{description}
\vspace{-0.025cm}
The top-K neighbors in the embedding space for a given entity may differ between KGEMs. However, this does not tell us much about why this is the case. We posit that studying the distribution of predicates in the K-hop subgraph centered around each neighboring entity can provide insights into the relevance of certain predicates for particular KGEMs. In other words, the subset of predicates that Rossi~\textit{et al.}~\cite{rossi2020relations} found out to influence entity representations and KGEM performance w.r.t. rank-based metrics might not just be dataset-dependent. Different KGEMs might also implicitly overweigh different subsets of predicates.

The main contributions of our work are summarized as follows.
\begin{itemize}
    \item We show that different KGEMs fulfill the \emph{KGE entity similarity assumption} only to a limited extent. Notably, even for a given KGEM, results can vary substantially on a per-class basis. Moreover, the semantics of classes is inequally captured by different KGEMs, thereby highlighting that different KGEMs expose different notions of similarity.
    \item We show that in most cases, performance in link prediction does not correlate with a model's adherence to the KGE entity similarity assumption. This demonstrates that rank-based metrics cannot be used as a reliable proxy for assessing the semantic consistency of the embedding space.
    \item We show that different KGEMs turn their attention to different predicate subsets for learning similar embeddings for related entities. This suggests that the notion of similarity in the embedding space is partially influenced by the predicate distribution in the close neighborhood around KG entities.
\end{itemize}

The remainder of the paper is structured as follows. Related work about KGEMs and their use in semantic-related tasks is presented in Section~\ref{sec:relatedwork}. Section~\ref{sec:approach} elaborates on our approach for measuring similarity. Section~\ref{sec:setting} details our experimental setting. Results are provided and discussed in Section~\ref{sec:results}. Lastly, Section~\ref{sec:conclusion} summarizes the key findings and outlines directions for future research.

\vspace{-0.25cm}
\section{Related Work}\label{sec:relatedwork}
\vspace{-0.15cm}

\sstitle{Knowledge graph embeddings.} KGEMs have garnered significant attention in recent years due to their capacity to represent structured knowledge within a continuous vector space. The seminal translational model TransE~\cite{transe} represents entities and relations as low-dimensional vectors and establishes the relationship between a head, a relation, and a tail through a translation operation in the embedding space. Subsequent KGEMs have primarily aimed to address its limitations and enhance the representational expressiveness of knowledge graph embeddings~\cite{wang2021survey}. Representative models are DistMult~\cite{distmult}, ComplEx~\cite{complex}, ConvE~\cite{conve}, and TuckER~\cite{tucker}. The embeddings learnt by these KGEMs have shown potential applications in tasks such as like link prediction~\cite{wang2021survey}, entity clustering~\cite{hubert2023maschine}, and node classification~\cite{hubert2023maschine}.

\sstitle{Using KGEs for semantic-related tasks.} As relational models, KGEMs are widely used for predicting links in KGs~\cite{wang2021survey,rossi2021survey}. However, the vector representations learnt by these models can also be used for other tasks~\cite{ji2021survey}. For example, pretrained language models with knowledge graphs have been used for Named Entity Recognition (NER)~\cite{sun2020ernie,liu2020kbert}. Other tasks aiming at discovering rich information about entities through their embeddings include entity typing~\cite{ma2016typing} and entity alignment~\cite{sun2020survey}. Many works also explore the use of KGEs for drug repurposing~\cite{sosa2020} and recommender systems~\cite{guo2022survey}. These works are based on the premise that the distance between entities in the embedding space should reflect their intrinsic similarity, and can be leveraged, \textit{e.g.} for recommending a common set of items to similar users~\cite{guo2022survey}.

\sstitle{Analzying the KGE entity similarity assumption.} While many of the approaches \emph{implicity} rely on the KGE entity similarity assumption to hold, there are only few works actually \emph{explicitly} validating this assumption. Portisch \textit{et al.}~\cite{portisch2022knowledge} provide anecdotic examples for a few embedding approaches, also suggesting that the underlying notions of similarity might differ, but do not conduct any formal evaluation. Jain \textit{et al.}~\cite{jain2021embeddings} analyze KGEMs on the basis of class assignments, showing that the original class assignments can only be reconstructed to a limited extent with classification and clustering methods, which rather questions the assumption of similar entities being close in the vector space. A similar study is conducted by Alshargi \textit{et al.}~\cite{alshargi2018concept2vec}, also concluding that ``the current quality of the embeddings for ontological concepts is not in a satisfactory state''. While Portisch \textit{et al.} provide anecdotic examples, and Jain \textit{et al.} and Alshargi \textit{et al.} only look at class assignments, this work is the first one to empirically study the relation of entity similarity and similarity in the KGE space. Moreover, our study is more fine-grained than the previous ones, which end at the class level: while those only inspect whether entities of the same class have similar embeddings (\textit{e.g.}, two movies are embedded closer than a movie and a person), we also consider similarity \emph{within} a class (\textit{e.g.}, analyze whether similar movies are embedded closer than less similar ones).

\vspace{-0.25cm}
\section{Approach}\label{sec:approach}
\vspace{-0.15cm}

In this section, we detail our proposal for quantifying the KGE entity similarity assumption. While similarity in embedding spaces is usually measured using cosine similarity, there is no uniform definition of entity similarity in KGs. In Section~\ref{sec:graph-similarity}, we elaborate on the metrics used to capture the notion of similarity between entities in the KG.
Section~\ref{sec:comparing-similarity} discusses how the aforementioned notions of similarity can be compared. It is worth noting that the line of research closest to ours is the one of Jain \textit{et al.}~\cite{jain2021embeddings}. Both~\cite{jain2021embeddings} and our work question common assumptions that are taken for granted in the KG community. However, \cite{jain2021embeddings} looks at the capability of KGEMs to learn the semantics of classes, and considers the specific tasks of entity classification and clustering. In contrast, we are concerned with similarity measures of different embedding models and how they reflect entity similarity. As such, even though our work fits within an existing literature on the concept of semantic capture in embeddings, to the best of our knowledge this paper is the first one to present an approach to thoroughly investigate the extent to which entity similarity in KGs is mirrored in the embedding space.

\vspace{-0.25cm}
\subsection{Towards a Graph-based Notion of Similarity}\label{sec:graph-similarity}
\vspace{-0.1cm}
In this section, we detail our attempt at capturing the notion of similarity in the original KG. Unlike the notion of similarity in the embedding space, which is typically measured as the cosine similarity of two vectors, it should be noted that graph-based similarity cannot be measured in a single, uniform way. As previsouly said, there is no universal definition for entity similarity in KGs. Multiple metrics and approaches can be used, and the choice between one or another largely depends on the particular aspect to be measured. In what follows, we intend to explain the rationale behind our modeling choices, and briefly mention alternatives that we ultimately discarded.

To measure the similarity between two entities $e_1$ and $e_2$ in one KG, we (1) determine the set of common statements about $e_1$ and $e_2$, \textit{i.e.}, relations to other entities that $e_1$ and $e_2$ have in common. While many entities will have such relations in common with central entities (\textit{i.e.}, those with many ingoing and outgoing edges), we also (2) need to make sure that centrality does not skew our similarity measure. Finally, (3) since an entity cannot be fully described by its immediate neighbors, indirect dependencies also need to be captured.

\begin{figure}[t]
    \centering
    \includegraphics[scale=0.7]{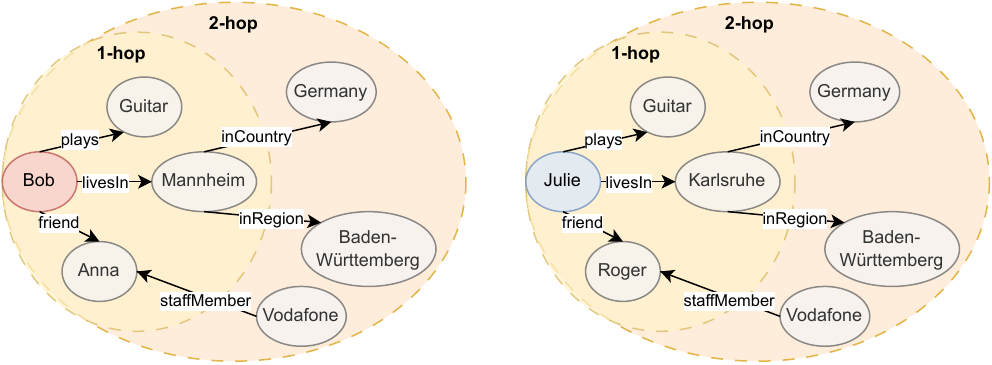}
    \caption{1-hop and 2-hop subgraphs for \texttt{Bob} and \texttt{Julie}.}
    \label{fig:example-subgraphs}
\end{figure}

To satisfy our first desideratum (1), we compare the subgraphs around $e_1$ and $e_2$. We experiment with the 1-hop only vs. 2-hop subgraph neighborhood. The latter option addresses desideratum (3), as we also consider indirect dependencies.
In Fig.~\ref{fig:example-subgraphs}, we give a concrete example of the 1-hop and 2-hop subgraphs for entities \texttt{Bob} ($e_1$) and \texttt{Julie} ($e_2$).\footnote{We consider both ingoing and outgoing edges to define those neighborhoods.}
The similarity of $e_1$ and $e_2$ can now be measured by the similarity of their respective subgraphs. Graph Edit Distance (GED)~\cite{sanfeliu1983} has been used for this intent. However, GED is NP-hard and thus computationally demanding, it comes with the need for arbitrarily defining weights for vertex and edge insertion/deletion/substitution, and we experimentally found it was sensitive to subgraphs' sizes, which is dentrimental to desideratum (2).
We considered other metrics such as Katz centrality and the common-neighbors metric. However, as the name suggests, Katz is a measure of centrality, not similarity. Besides, it takes into account all the paths between two entities and is therefore sensitive to the absolute number of paths. Common neighbors does not consider predicates and is only suited for unlabeled graphs.\footnote{Given, e.g., \texttt{(SAP, headquarter, Germany), (BOSCH, headquarter, Germany), (Berlin, capitalOf, Germany)}, with common neighbors, \texttt{SAP}, \texttt{BOSCH}, and \texttt{Berlin} would be equally similar.}

In our experiments, we use the Jaccard coefficient is used to measure the overlap in the 1-hop and 2-hop subgraphs of entities. In particular, its value denotes how similar the respective subgraphs of two entities are, which gives us insight into how much two entities are related based on graph-based information.
The central entity ($e_1$ or $e_2$) is replaced by a unique token, \textit{e.g.} \texttt{:dummy}. Triples forming the 1-hop neighborhood are extracted as is, while for the 2-hop paths the intermediate entity is ignored (written as $\Box$ below). 

More formally:
let $\mathcal{T}_1(A)$ be the set of all triples $\{(A, p, o) \cup (s, p, A)\}$, where $A$ is replaced by a dummy identifier. Then, $J_1(e_1,e_2)$ is the Jaccard overlap of $\mathcal{T}_1(e_1)$ and $\mathcal{T}_1(e_2)$, \textit{i.e.}, $\frac{|\mathcal{T}_1(e_1) \cap \mathcal{T}_1(e_2)|}{|\mathcal{T}_1(e_1) \cup \mathcal{T}_1(e_2)|}$.

Let $\mathcal{T}_2(A)$ be the set of all paths $\{(A, p1, \Box, p2, o) \cup (s, p1, \Box, p2, A)\}$, where $A$ is replaced by a dummy identifier. Then $J_2(e_1,e_2)$ is the Jaccard overlap of $\mathcal{T}_2(e_1)$ and $\mathcal{T}_2(e_2)$, \textit{i.e.}, $\frac{|\mathcal{T}_2(e_1) \cap \mathcal{T}_2(e_2)|}{|\mathcal{T}_2(e_1) \cup \mathcal{T}_2(e_2)|}$.

\vspace{0.15cm}
Following Fig.~\ref{fig:example-subgraphs}, triples in the two 1-hop subgraphs are:
{\footnotesize
\begin{align*}
\begin{array}{c|c}
    \text{Bob} & \text{Julie} \\
    \midrule
    \texttt{(:dummy, plays, Guitar)} & \texttt{(:dummy, plays, Guitar)} \\
    \texttt{(:dummy, livesIn, Mannheim)} \quad & \quad \texttt{(:dummy, livesIn, Karlsruhe)} \\
    \texttt{(:dummy, friend, Anna)} & \texttt{(:dummy, friend, Roger)}
\end{array}
\end{align*}
}

When considering 2-hop information, we also consider the following paths which are shared between Bob's and Julie's subgraphs (reverse edges included):
{\footnotesize
\begin{align*}
    \texttt{(:dummy, livesIn, $\Box$, inCountry, Germany)} \\
    \texttt{(:dummy, livesIn, $\Box$, inRegion, Baden-Württemberg)} \\
    \texttt{(:dummy, friend, $\Box$, staffMember$^{-1}$, Vodafone)}
\end{align*}
}

The similarity of \texttt{Bob} and \texttt{Julie} considering the 1-hop subgraphs only is $J_1(Bob, Julie) = 0.2$ (since they share one out of five triples). Considering the 2-hop neighborhood graphs: $J_2(Bob, Julie) = 0.5$ (since they share three out of six two-hop paths).
For instance, we consider \texttt{(:dummy, livesIn, $\Box$, inCountry, Germany)} as the full path.
This allows for capturing indirect dependencies, such as \emph{both Julie and Bob live in a city in Germany}, or \emph{both Julie and Bob know someone who works for Vodafone}.
This approach better fits the general and commonsense idea that two entities should already be similar if they share relations to entities which are similar themselves according to their respecitve 1-hop subgraphs. Moreover, keeping the intermediate entity in the path would overweigh information already counted as part of the 1-hop subgraph, especially if the intermediate entity were a high degree entity (desideratum (3)). For any entity in the 1-hop neighborhood of a central entity, its own 1-hop neighborhood is known and fixed. Consequently, whenever there is a match between triples in the 1-hop neighborhood of two central entities being compared, there would be as many matches in their respective 2-hop subgraphs as there are relations connecting their shared 1-hop neighbors. For example, if \texttt{Bob} and \texttt{Julie} were living in the same city, \textit{e.g.} Mannheim, this information would be counted multiple times -- in \texttt{(:dummy, livesIn, Mannheim, inCountry, Germany)}, \texttt{(:dummy, livesIn, Mannheim, inRegion, Baden-Württemberg)}, etc.. The resulting relatedness between \texttt{Bob} and \texttt{Julie} would thus be overly influenced by that single fact. Similarly, if they were living in distinct cities -- which is the case in our example -- their relatedness score would be unduly penalized.

\vspace{-0.25cm}
\subsection{Embedding vs. Graph: a Different Notion of Similarity?}\label{sec:comparing-similarity}
\vspace{-0.15cm}
Based on a given entity $e$, it is possible to get its $N$ closest neighbors in the KG (using the $J_1$ and $J_2$ metrics defined above) and in the embedding space (using cosine similarity).
To assess if the two notions of similarity are different, we need to measure how much the two lists of closests neighbors between these two approaches actually overlap.

A common measure to compare two ranked lists is Kendall's Tau~\cite{kendall1938}. However, it suffers from two important caveats: the two lists should be of the same size and they should contain the same set of items. Considering our experimental purposes, the latter limitation is undesirable, as there is no reason why these two lists (of the top $N$ similar entities according to J$_1$ or J$_2$, and in the embedding space) should share the same set of entities. 

Consequently, we use the Rank-Biased Overlap (RBO)~\cite{webber2010rbo} to compare the similarity of two ranked lists. Unlike Kendall's Tau which is correlation-based, RBO is intersection-based. Most importantly, RBO can handle lists containing different items. It also allows for the weighting of rankings, giving higher importance to items at the top of the lists through tweaking the \emph{persistence} parameter $p$. A small value for $p$ will only consider the first few items, whereas a larger value will encompass more items. However, we experimentally found that results are very sensitive to the choice of $p$. Besides, when comparing two ranked lists, we want to consider all of their items. Consequently, we stick to the default parameter strategy $p = 1$ as proposed in an open-source implementation of RBO\footnote{\url{https://github.com/changyaochen/rbo/}} that we used in our experiments. Formally, RBO is expressed as follows.

Let $S$ and $T$ be two ranking lists, and let $S_{i}$ (resp. $T_{i}$) be the element at rank $i$ in list $S$ (resp. $T$). Then, $S_{c:d}$ (resp. $T_{c:d}$) denotes the set of the elements from position $c$ to position $d$ in list $S$ (resp. $T$). At each depth $d$, the intersection of lists $S$ and $T$ to depth $d$ is defined as $I_{S,T,d} = S_{1:d} \cap T_{1:d}$, and their overlap up to depth $d$ is the size of the intersection, \textit{i.e.} $|I_{S,T,d}|$. RBO relies on the notion of \emph{agreement} between $S$ and $T$ to depth $d$:
\begin{align}
A_{S,T,d} = \frac{|I_{S,T,d}|}{d}
\end{align}
In the case whre the persistence parameter $p = 1$, the RBO formula is expressed as follows:
\begin{align}
RBO(S, T, k) = \frac{1}{k} \sum_{d=1}^{k} A_{S,T,d}
\end{align}
where $k$ is the size of the two lists $S$ and $T$ being compared. For the case where $p \neq 1$, we refer the reader to~\cite{webber2010rbo}.

\begin{figure}
    \centering
    \includegraphics[scale=0.55]{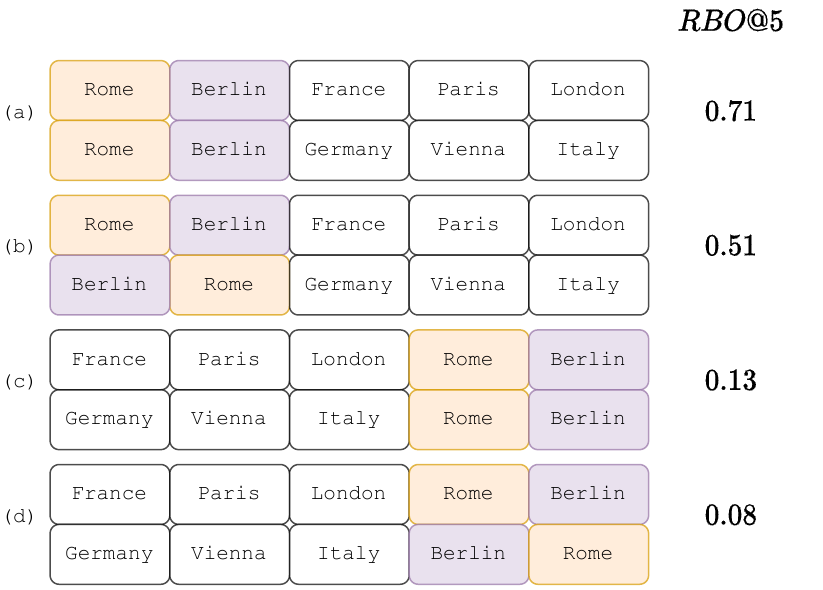}
    \caption{Illustration of RBO@5 values in four cases.}
    \label{fig:example-rbo}
\end{figure}

In Fig.~\ref{fig:example-rbo}, we illustrate how RBO@5 is calculated for two lists of top-5 neighbors, where one relies on similarity in the embedding space whereas the other one depends on the Jaccard overlap between entities' subgraphs. We see that (i) RBO can be calculated on lists with different sets of entities, (ii) RBO weighs more matches that occur at the top of the lists, and (iii) RBO is sensitive to the order of matching entities in the respective lists.

\vspace{-0.25cm}
\section{Experiments}\label{sec:setting}
In this section, we detail our experimental setting, \textit{i.e.} the KGEMs (Section~\ref{sec:kgems}) and datasets (Section~\ref{sec:datasets}) used in the experiments.
\vspace{-0.25cm}
\subsection{Knowledge Graph Embedding Models}\label{sec:kgems}
\vspace{-0.15cm}
In our experiments, we use seven popular KGEMs from different families of models: geometric-based (TransE~\cite{transe}, TransD~\cite{transd}, BoxE~\cite{boxe}), multiplicative (RESCAL \cite{rescal}, DistMult~\cite{distmult}, TuckER~\cite{tucker}), and convolutional-based (ConvE~\cite{conve}) models. Other models have been considered, \textit{e.g.} ComplEx~\cite{complex} and RotatE~\cite{rotate}. However, these models generate complex-valued embeddings, which requires a different approach than using simple cosine similarity. We trained these KGEMs using PyKEEN\footnote{\url{https://github.com/pykeen/pykeen/}} with the provided configuration files, when available. For those datasets with no reported best hyperparameters, we used the hyperparameters reported in the original paper (\textit{e.g.} for YAGO4-19K, with best hyperparameters found in~\cite{hubert2023sematk}) or performed manual hyperparameter search and kept the sets of hyperparameters leading to the best results on the validation sets (\textit{e.g.} for AIFB). We also trained RDF2Vec~\cite{ristoski2016rdf2vec}, which is a versatile embedding approach that can be adapted for different downstream applications and that relies on path-based information to encode entities (only). Even though RDF2Vec is not specifically designed to handle the link prediction task and is not evaluated w.r.t. it, its entity embeddings are expected to reflect their similarity or relatedness in the KG to some extent~\cite{portisch2022knowledge}. We used the implementation provided by pyRDF2Vec\footnote{\url{https://github.com/IBCNServices/pyRDF2Vec/}} with the default hyperparameters reported in~\cite{portisch2022knowledge}: embeddings of dimension $200$, with $2,000$ walks maximum, a depth of $4$, a window size of $5$, and $25$
epochs for training word2vec~\cite{mikolov2013} with the continuous skip-gram architecture. 

\vspace{-0.5cm}
\subsection{Datasets}\label{sec:datasets}
\vspace{-0.25cm}
Since we also want to analyze results on a per-class basis, we consider only benchmark datasets which also come with a schema of multiple classes and relations. We use the following datasets in our evaluation: AIFB~\cite{aifb}, Codex-S and Codex-M~\cite{codex}, DBpedia50~\cite{shi2018conmask}, FB15K-237~\cite{toutanova}, and YAGO4-19K~\cite{hubert2023sematk}. Entity types are not directly available in the original repository of DBpedia50\footnote{\url{https://github.com/bxshi/ConMask/}} and FB15k-237\footnote{\url{https://www.microsoft.com/en-us/download/details.aspx?id=52312/}}. However, we ran SPARQL queries against DBpedia to get entity types for DBpedia50. The resulting class hierarchy is a subset of the DBpedia ontology, with a maximum depth of $8$, an average fan-out (branching factor) of $3.40$, and each entity being typed with an average of $4.55$ classes. For FB15K-237, we reused the entity-typed version presented in~\cite{hubert2023sematk}.
Table~\ref{tab:datasets} provides finer-grained statistics for these datasets.
\begin{table}[t]
\footnotesize
\centering
\caption{Datasets used in the experiments. Column header from left to right: number of entities, relations (predicates), classes, train triples, validation triples, and test triples.}\label{tab:datasets}
\setlength{\tabcolsep}{10pt}
\begin{tabular}{lrrrrrr}
\toprule
Dataset & \multicolumn{1}{c}{$|\mathcal{E}|$} & \multicolumn{1}{c}{$|\mathcal{R}|$} &
\multicolumn{1}{c}{$|\mathcal{C}|$} & \multicolumn{1}{c}{$|\mathcal{T}_{train}|$} & \multicolumn{1}{c}{$|\mathcal{T}_{valid}|$} & \multicolumn{1}{c}{$|\mathcal{T}_{test}|$}\\
\midrule
AIFB & $2,389$ & $16$ & $18$ & $14,170$ & $745$ & $785$  \\
Codex-S & $2,034$ & $42$ & $502$ & $32,888$ & $1,827$ & $1,827$  \\
Codex-M & $17,050$ & $51$ & $1,503$ & $185,584$ & $10,310$ & $10,311$  \\
DBpedia50  & $24,624$ & $351$ & $285$ & $32,388$ & $123$ & $2,098$ \\
FB15k237 & $14,541$ & $237$ & $643$ & $272,115$ & $17,535$ & $20,466$ \\
YAGO4-19k & $18,960$ & $74$ & $1,232$ & $27,447$ & $485$ & $463$  \\
\bottomrule
\end{tabular}
\end{table}

\section{Results and Discussion}\label{sec:results}
\subsection{Different Notions of Similarity (RQ1)}
Table~\ref{tab:expe1-results} reports results w.r.t. rank-based metrics (MRR, Hits@K) and RBO@K for all the KGEMs and datasets considered in this work. Similar to MRR and Hits@K, RBO is computed for all entities in a KG and the average is reported.

When retaining 1-hop subgraphs (from R1@3 to R1@100 in Table~\ref{tab:expe1-results}), in most cases TuckER has the highest RBO values. In other words, it appears to fulfill the KGE entity similarity assumption best, having the highest tendency to position similar entities closer in the vector space.
When moving to 2-hop subgraphs, DistMult and TuckER are the models with the best alignment capabilities (Table~\ref{tab:expe1-results}).
BoxE, RDF2Vec, RESCAL, and TransD fare worse than other models. The substantially lower RBO values for RDF2Vec can be related to the fact that the vector distance in RDF2vec space mixes similarity and relatedness, where the latter does not show a strong overlap in common paths~\cite{portisch2022knowledge}. For example, it will position \emph{USA} close to \emph{Washington D.C.}, but they have only few common paths starting/ending in the respective entities.

A crucial observation from our study is that the behavior of KGEMs varies not only across datasets but also across different classes within these datasets.
Table~\ref{tab:examples-rbo} illustrates this insight on classes that were selected to provide a more intuitive understanding of how different KGEMs are differently able to capture similarities between instances of the same class. In particular, we clearly see that on YAGO4-19K, TransE captures the notion of \texttt{SpanishMunicipality} quite well compared to other models, but does not do well for \texttt{MusicPlaylist} entities (highlighted in blue).
This means that different models are better suited for capturing the nuances of specific classes. Jain \textit{et al.}~\cite{jain2021embeddings} demonstrated that semantic representation in embeddings is not consistent across all KG entities, but is restricted to a small subset of them. Our results point in the same direction and suggest that KGEMs are not equipped with consistent capabilities to learn similar embeddings for entities within the same class.
This problem appears at two different levels: First, we noted a general misalignment in the representation of certain classes across various datasets, regardless of the KGEM used. Second, there is noticeable inconsistency within specific classes, where entities might align well with their graph-based proximity in one KGEM but not in another.

\begin{figure}[t]
    \centering
    \includegraphics[scale=0.4]{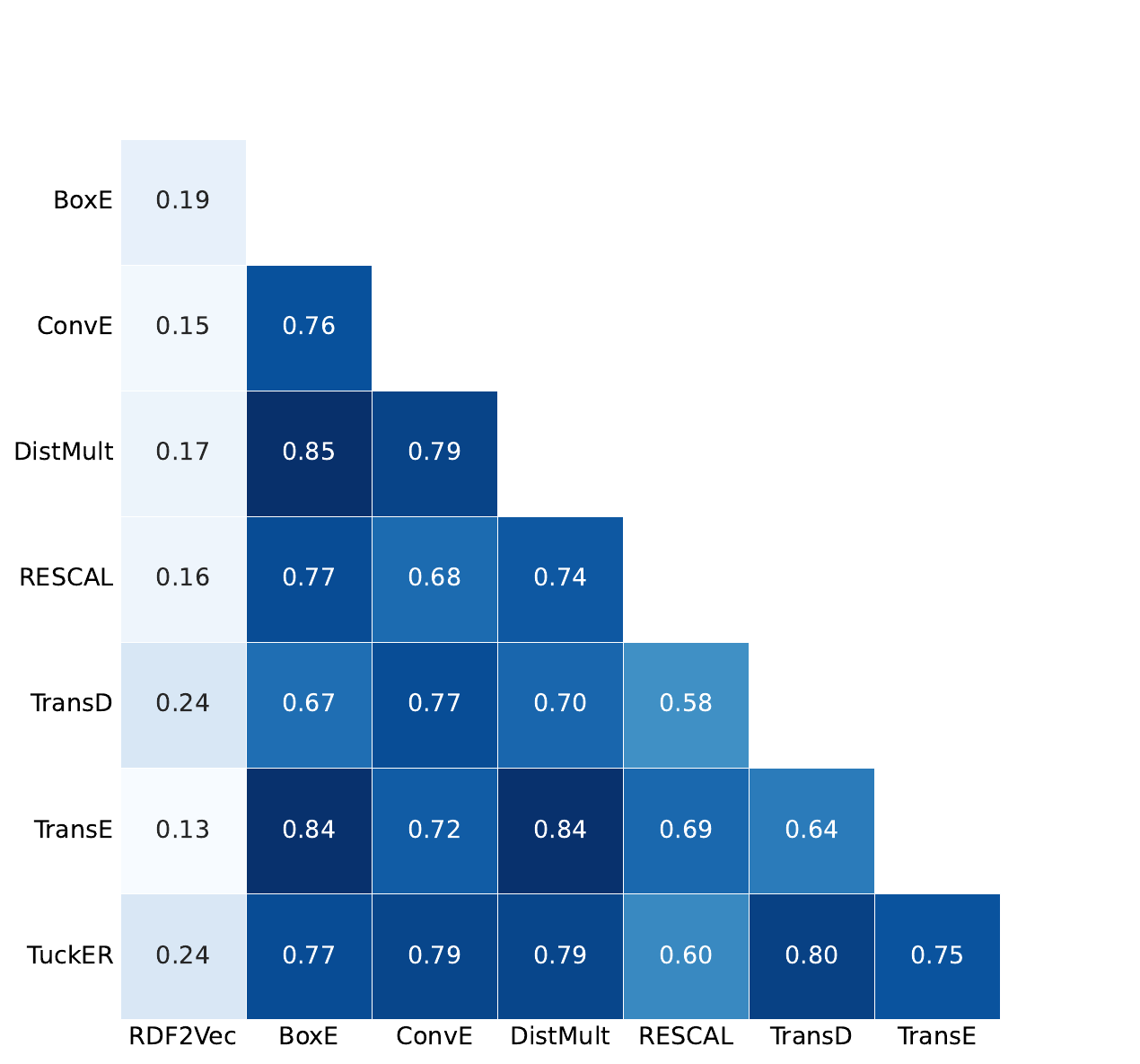}
    \caption{Rank correlations between KGEMs averaged over all datasets and all RBO metrics ($K={3,10,100}$).}
    \label{fig:correl_matrix}
\end{figure}

To generalize this observation, we computed correlation matrices for RBO values across models and datasets, focusing on class-specific performance.
Our analysis, averaging these matrices, gives us a broader view of how different KGEMs perceive similarity on a per-class basis. Fig.~\ref{fig:correl_matrix} shows the rank correlations between models when considering 2-hop subgraphs, averaged over all datasets and RBO values. Notably, RDF2Vec stands out as it seems to capture a distinct notion of similarity compared to other KGEMs (cf.~\cite{portisch2022knowledge}). It demonstrates that RDF2Vec fares better on different classes than other KGEMs. 
We also observe a high degree of variation between models of the same family: TransE and TransD expose a different notion of similarity (0.64), as RESCAL and TuckER do (0.6). Strong rank correlations are observed between DistMult and BoxE (0.85), DistMult and TransE (0.84), and BoxE and TransE (0.84) (Fig.~\ref{fig:correl_matrix}). This demonstrates that while some KGEMs exhibit a close conceptualization of entity similarity, this is not universally the case. Therefore, \textbf{RQ1} can be answered as follows: proximity in the embedding space does \emph{not} consistently align with the notion of entity similarity in the KG, since this property substantially differs between models and datasets. Consequently, careful consideration is necessary when using KGEs for drawing conclusions about similar entities (\textit{e.g.} for recommending items to similar users). 

\begin{table}[t]
\centering
\caption{Rank-based and RBO results. As RDF2Vec is not designed for link prediction, it is not evaluated w.r.t. rank-based metrics. H@3 and H@10 stand for Hits@3 and Hits@10, respectively. RBO is abbreviated as R, while the number directly following it denotes whether 1-hop subgraphs or 2-hop subgraphs are considered. Bold fonts indicate which KGEM performs best for a given configuration (dataset-metric). Underlined results denote the second-best performing model.}\label{tab:expe1-results}
\resizebox{\textwidth}{!}{%
\setlength{\tabcolsep}{1.75pt}
\begin{tabular}{cccccccccccccccccccccc}
\toprule
&& \multicolumn{9}{c}{FB15K-237} && \multicolumn{9}{c}{YAGO4-19K}\\
\cmidrule{3-11} \cmidrule{13-21}
&& MRR & H@3 & H@10 & R1@3 & R1@10 & R1@100 & R2@3 & R2@10 & R2@100 && MRR & H@3 & H@10 & R1@3 & R1@10 & R1@100 & R2@3 & R2@10 & R2@100\\
\midrule
RDF2Vec & &  \dashed & \dashed & \dashed  & 0.023 & 0.033 & 0.063 & 0.014 & 0.021 & 0.054 &&  \dashed & \dashed & \dashed  & 0.126 & 0.154 & 0.177 & 0.164 & 0.170 & 0.183 \\ \midrule
TransE & & 0.240 & 0.264 & 0.404 & 0.132 & 0.178 & 0.228 & 0.035 & 0.050 & 0.108 && 0.762 & 0.836 & 0.895 & 0.247 & 0.301 & 0.360 & 0.368 & 0.362 & \textbf{0.260} \\ \midrule
TransD & & 0.184 & 0.205 & 0.378 & 0.176 & 0.220 & 0.259 & 0.065 & 0.083 & 0.135 && 0.763 & 0.895 & 0.915 & 0.197 & 0.252 & 0.365 & 0.181 & 0.237 & 0.137 \\ \midrule
DistMult & & 0.226 & 0.247 & 0.392 & 0.194 & 0.245 & 0.343 & \textbf{0.103} & \textbf{0.131} & \textbf{0.222} && 0.809 & 0.838 & 0.870 & 0.310 & 0.350 & \underline{0.407} & \underline{0.378} & \textbf{0.372} & 0.225 \\ \midrule
RESCAL & & 0.279 & 0.305 & 0.447 & 0.239 & 0.269 & 0.284 & 0.064 & 0.081 & 0.138 && 0.676 & 0.692 & 0.754 & 0.176 & 0.215 & 0.176 & 0.298 & 0.271 & 0.103 \\ \midrule
TuckER & & \textbf{0.341} & \textbf{0.373} & \textbf{0.516} & \underline{0.321} & \underline{0.364} & \textbf{0.411} & \underline{0.099} & \underline{0.121} & \underline{0.205} && 0.897 & 0.901 & 0.910 & \textbf{0.362} & \textbf{0.461} & \textbf{0.498} & \textbf{0.385} & \underline{0.362} & \underline{0.228} \\ \midrule
ConvE & & 0.300 & 0.327 & 0.474 & \textbf{0.341} & \textbf{0.370} & \underline{0.357} & 0.083 & 0.102 & 0.153 && \textbf{0.905} & \textbf{0.907} & \textbf{0.916} & \underline{0.328} & \underline{0.368} & 0.365 & 0.336 & 0.313 & 0.168 \\ \midrule
BoxE & & 0.299 & 0.326 & 0.477 & 0.221 & 0.261 & 0.313 & 0.060 & 0.081 & 0.166 && 0.895 & 0.902 & 0.914 & 0.136 & 0.146 & 0.113 & 0.244 & 0.231 & 0.096 \\
\toprule
&& \multicolumn{9}{c}{Codex-S} && \multicolumn{9}{c}{Codex-M}\\
\cmidrule{3-11} \cmidrule{13-21}
&& MRR & H@3 & H@10 & R1@3 & R1@10 & R1@100 & R2@3 & R2@10 & R2@100 && MRR & H@3 & H@10 & R1@3 & R1@10 & R1@100 & R2@3 & R2@10 & R2@100\\
\midrule
RDF2Vec & & \dashed & \dashed & \dashed & 0.030 & 0.043 &  0.096 & 0.021 & 0.036 & 0.099 &&  \dashed & \dashed & \dashed  & 0.013 & 0.019 & 0.042 & 0.011 & 0.017 & 0.041 \\ \midrule
TransE & & 0.293 & 0.331 & 0.526 & 0.201 & 0.266 & 0.451 & \underline{0.154} & \textbf{0.210} & \textbf{0.345} && 0.227 & 0.260 & 0.371 & 0.082 & 0.101 & 0.153 & 0.058 & 0.076 & 0.105 \\ \midrule
TransD & & 0.226 & 0.294 & 0.515 & 0.252 & 0.322 & 0.462 & 0.136 & 0.178 & 0.297 && 0.215 & 0.281 & 0.420 & 0.155 & 0.188 & 0.263 & 0.055 & 0.070 & 0.117 \\ \midrule
DistMult & & 0.261 & 0.298 & 0.409 & \underline{0.452} & \textbf{0.511} & \textbf{0.616} & \textbf{0.163} & 0.205 & \underline{0.339} && 0.209 & 0.233 & 0.356 & 0.173 & 0.212 & 0.312 & \textbf{0.144} & \textbf{0.178} & \textbf{0.256} \\ \midrule
RESCAL & & 0.283 & 0.307 & 0.475 & 0.135 & 0.193 & 0.336 & 0.132 & 0.176 & 0.274 && 0.149 & 0.159 & 0.253 & 0.067 & 0.084 & 0.115 & 0.025 & 0.036 & 0.057 \\ \midrule
TuckER & & 0.393 & 0.442 & 0.618 & \textbf{0.461} & \textbf{0.511} & \underline{0.615} & \underline{0.154} & \underline{0.209} & \underline{0.339} && 0.282 & 0.310 & 0.424 & 0.198 & 0.239 & 0.328 & \underline{0.120} & \underline{0.153} & \underline{0.222} \\ \midrule
ConvE & & \textbf{0.414} & \textbf{0.466} & 0.611 & 0.450 & 0.495 & 0.574 & 0.146 & 0.195 & 0.306 && \textbf{0.290} & \textbf{0.319} & 0.425 & \textbf{0.315} & \textbf{0.343} & \textbf{0.384} & 0.098 & 0.119 & 0.158 \\ \midrule
BoxE & & 0.398 & 0.453 & \textbf{0.622} & 0.312 & 0.366 & 0.486 & 0.118 & 0.179 & 0.291 && 0.290 & 0.318 & \textbf{0.431} & 0.241 & \underline{0.272} & \underline{0.334} & 0.084 & 0.106 & 0.152 \\
\toprule
&& \multicolumn{9}{c}{AIFB} && \multicolumn{9}{c}{DBpedia50}\\
\cmidrule{3-11} \cmidrule{13-21}
&& MRR & H@3 & H@10 & R1@3 & R1@10 & R1@100 & R2@3 & R2@10 & R2@100 && MRR & H@3 & H@10 & R1@3 & R1@10 & R1@100 & R2@3 & R2@10 & R2@100\\
\midrule
RDF2Vec & &  \dashed & \dashed & \dashed  & 0.089  & 0.131 & 0.223 & 0.137 & 0.174 & 0.210 &&  \dashed & \dashed & \dashed  & 0.04 & 0.053 & 0.179 & 0.059 & 0.082 & 0.189 \\ \midrule
TransE & & 0.469 & 0.524 & 0.699 & 0.413 & 0.456 & 0.571 & 0.304 & 0.349 & \underline{0.455} && 0.397 & \textbf{0.490} & \textbf{0.567} & 0.109 & 0.123 & 0.297 & \underline{0.214} & 0.246 & 0.282 \\ \midrule
TransD & & 0.472 & 0.689 & 0.817 & 0.472 & 0.507 & \underline{0.617} & 0.305 & 0.330 & 0.382 && 0.260 & 0.376 & 0.442 & 0.111 & 0.136 & 0.318 & 0.126 & 0.187 & 0.244 \\ \midrule
DistMult & & 0.500 & 0.568 & 0.763 & 0.423 & 0.463 & 0.588 & \underline{0.358} & \textbf{0.406} & \textbf{0.495} && 0.384 & 0.423 & 0.478 & 0.131 & 0.144 & 0.348 & \textbf{0.241} & \textbf{0.275} & \underline{0.299} \\ \midrule
RESCAL & & 0.440 & 0.486 & 0.628 & 0.345 & 0.387 & 0.454 & 0.245 & 0.273 & 0.322 && 0.225 & 0.240 & 0.295 & 0.063 & 0.079 & 0.276 & 0.142 & 0.171 & 0.197 \\ \midrule
TuckER & & 0.797 & 0.824 & 0.888 & \textbf{0.693} & \textbf{0.712} & \textbf{0.726} & \textbf{0.375} & 0.380 & 0.394 && 0.424 & 0.442 & 0.502 & \textbf{0.151} & \textbf{0.166} & \textbf{0.418} & 0.235 & \underline{0.271} & \textbf{0.305} \\ \midrule
ConvE & & 0.779 & 0.799 & 0.870 & 0.454 & 0.505 & 0.567 & 0.263 & 0.292 & 0.321 && 0.434 & 0.456 & 0.536 & \underline{0.137} & 0.146 & 0.324 & 0.201 & 0.228 & 0.253 \\ \midrule
BoxE & & \textbf{0.806} & \textbf{0.841} & \textbf{0.904} & \underline{0.570} & \underline{0.587} & 0.614 & 0.343 & \underline{0.374} & 0.437 && \textbf{0.456} & 0.480 & 0.556 & \underline{0.137} & \underline{0.147} & \underline{0.351} & \underline{0.214} & 0.247 & 0.278 \\
\bottomrule
\end{tabular}
}
\end{table}

\begin{table}[t]
\centering
\caption{RBO@10 values per model for selected classes. Numbers in parenthesis indicate the rank of a class for a given model.}\label{tab:examples-rbo}
\footnotesize
\resizebox{\textwidth}{!}{%
\setlength{\tabcolsep}{3.5pt}
\begin{tabular}{llllllllll}
\toprule
Dataset & Class & RDF2Vec & TransE & TransD & BoxE & RESCAL & DistMult & Tucker & ConvE\\
\midrule
\multirow{2}{*}{AIFB} & \texttt{Book} & 0.383 (1) & 0.560 (1) & 0.314 (7) & 0.551 (1) & 0.613 (1) & 0.621 (1) & 0.374 (5) & 0.214 (13)\\
& \texttt{inProceedings} & 0.048 (13) & 0.419 (3) & 0.227 (11) & 0.461 (3) & 0.412 (2) & 0.453 (3) & 0.332 (11) & 0.326 (3)\\
\midrule
\multirow{2}{*}{Codex-S} & \texttt{Republic} & 0.000 (26) & 0.556 (3) & 0.236 (12) & 0.463 (5) & 0.363 (3) & 0.339 (10) & 0.399 (9) & 0.311 (12)\\
& \texttt{NationalAcademy} & 0.000 (27) & 0.558 (2) & 0.217 (14) & 0.446 (8) & 0.383 (2) & 0.323 (11) & 0.366 (12) & 0.301 (14)\\
\midrule
\multirow{2}{*}{Codex-M} & \texttt{UrbanDistrict} & 0.002 (66) & 0.538 (9) & 0.178 (40) & 0.482 (11) & 0.057 (66) & 0.494 (14) & 0.504 (8) & 0.327 (12)\\
& \texttt{CollegeTown} & 0.000 (88) & 0.615 (2) & 0.205 (30) & 0.653 (2) & 0.133 (30) & 0.670 (1) & 0.640 (2) & 0.337 (27)\\
\midrule
\multirow{2}{*}{DBpedia50} & \texttt{Album} & 0.134 (52) & 0.336 (33) & 0.270 (23) & 0.351 (30) & 0.300 (9) & 0.310 (42) & 0.387 (27) & 0.315 (29)\\
& \texttt{EthnicGroup} & 0.028 (136) & 0.469 (9) & 0.227 (36) & 0.452 (9) & 0.400 (1) & 0.458 (9) & 0.434 (16) & 0.388 (12)\\
\midrule
\multirow{2}{*}{FB15k237} & \texttt{Periodicals} & 0.002 (414) & 0.115 (109) & 0.255 (64) & 0.150 (163) & 0.124 (181) & 0.139 (280) & 0.590 (11) & 0.638 (3)\\
& \texttt{BaseballPlayer} & 0.001 (420) & 0.146 (78) & 0.336 (31) & 0.236 (93) & 0.143 (160) & 0.277 (104) & 0.631 (4) & 0.598 (12)\\
\midrule
\multirow{2}{*}{YAGO-19k} & \texttt{MusicPlaylist} & 0.233 (35) & \textbf{\blue{0.325 (94)}} & 0.322 (37) & 0.411 (10) & 0.350 (41) & 0.366 (83) & 0.627 (4) & 0.554 (7)\\
& \texttt{SpanishMunicipality} & 0.226 (40) & \textbf{\blue{0.636 (1)}} & 0.460 (4) & 0.327 (33) & 0.236 (95) & 0.531 (20) & 0.551 (20) & 0.505 (17)\\
\bottomrule
\end{tabular}
}
\end{table}

\vspace{-0.15cm}
\subsection{Correlation between Rank-based Metrics and RBO (RQ2)}
\vspace{-0.05cm}

\begin{figure}[t]
    \centering
    \includegraphics[scale=0.45]{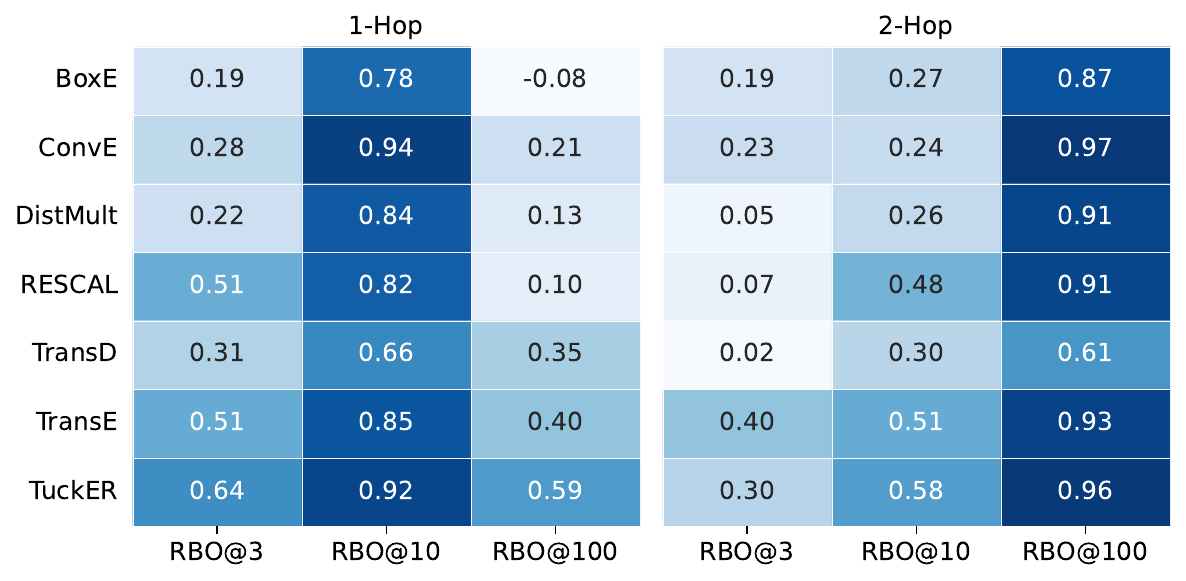}
    \caption{Correlations between MRR and RBO values with 1-hop and 2-hop subgraphs.}
    \label{fig:mrr-rbo}
\end{figure}

As previously mentioned, some KGEMs fare better than others in terms of RBO values, \textit{e.g.} DistMult and TuckER. 
It is important to note that RBO values are designed to measure how well a KGEM entity embedding aligns with the concept of proximity within the original KG. This goal is distinct from the aim of rank-based metrics, which assess how effectively a KGEM assigns higher plausibility scores to ground-truth triples.
In what follows, we aim study the relationship between these two types of metrics: how does KGEM performance in rank-based metrics correlate with RBO values?

We analyzed this using Pearson's correlation coefficient, comparing MRR and RBO values for 1-hop and 2-hop subgraphs across our 6 datasets, as depicted in Fig.~\ref{fig:mrr-rbo}. 
The results revealed notable disparities among the models. For instance, TransE and TuckER consistently show moderate to high correlation between MRR and RBO values. This implies that for these models, higher plausibility scoring of triples often aligns with better representation of entity proximity in the KG. However, this trend is not uniform across all models, and we do not observe general trends across families of models: while TransE and TransD are both translational models, correlation results are quite different. 

From a coarse-grained viewpoint, we note that MRR \emph{generally} does not correlate well with RBO values. Two specific cases are that MRR is heavily (and positively) correlated with RBO@10 (1-hop) and RBO@100 (2-hop), which means that MRR can be seen as a reliable proxy measure for those two metrics (and vice versa). However, it still remains to be explained why the correlation between MRR and RBO@10 (1-hop) is so high, while a substantial drop in correlation between MRR and RBO@100 (1-hop) is observed. Another result to be assessed is why MRR correlates more with RBO@K with lower $K$ values when considering 1-hop subgraph and higher $K$ values when considering 2-hop subgraphs. Therefore, \textbf{RQ2} cannot be answered in a conclusive way: while in the general case, the correlation between performance in grouping similar entities and link prediction performance is moderate at best, some metrics show a higher correlation. This has a severe practical implication: the commonly observed practice to pick a KGE model which is good at link prediction for a task heavily relying on entity similarity has to be considered a suboptimal strategy.

\vspace{-0.25cm}
\subsection{Analyzing Predicate Importance (RQ3)}
\vspace{-0.15cm}
We previously highlighted that different KGEMs are equipped with different notions of similarity in their respective embedding spaces. We also demonstrated that the adherence to the KGE entity similarity assumption can vary a lot between classes: for a given combination of dataset and model, how much the top-K neighbors of an entity in the embedding space align with its top-K neighbors in the graph largely depends on which class this entity belongs to.

To examine this behavior further, we pick random classes and count the predicates in the common triples of the top-K neighbors (in the embedding space) for entities of this class differs between models. 

\begin{table}[h!]
\centering
\caption{Top 5 predicates for entities of classes \texttt{MusicComposition} and \texttt{MusicGroup} on YAGO4-19K, along with their importance score.}\label{tab:expe3-results}
\resizebox{\textwidth}{!}{%
\setlength{\tabcolsep}{3pt}
\begin{tabular}{crrrrrr}
\toprule
&& \multicolumn{2}{c}{\texttt{MusicComposition}} && \multicolumn{2}{c}{\texttt{MusicGroup}} \\
\cmidrule(lr){3-4} \cmidrule(lr){6-7}
&& K=3 & K=10 && K=3 & K=10 \\
\midrule
RDF2Vec
&& isPartOf (0.52) & isPartOf (0.46) && \blue{\textbf{genre (0.53)}} & \blue{\textbf{genre (0.60)}} \\
&& inLanguage (0.18) & inLanguage (0.30) && \blue{\textbf{foundingLocation (0.38)}} & \blue{\textbf{foundingLocation (0.32)}} \\
&& hasPart (0.17) & hasPart (0.15) && \blue{\textbf{award (0.10)}} & \blue{\textbf{award (0.07)}} \\
&& composer (0.09) & composer (0.07) && memberOf (0.00) & memberOf (0.01) \\
&& lyricist (0.02) & lyricist (0.02) && byArtist (0.00) & byArtist (0.00) \\
\midrule
TransE
&& isPartOf (0.40) & isPartOf (0.36) && genre (0.79) & genre (0.85) \\
&& hasPart (0.28) & hasPart (0.27) && foundingLocation (0.17) & foundingLocation (0.12) \\
&& inLanguage (0.16) & inLanguage (0.26) && award (0.04) & award (0.02) \\
&& composer (0.09) & composer (0.06) && memberOf (0.00) & memberOf (0.00) \\
&& citation (0.05) & citation (0.04) && byArtist (0.00) & knowsLanguage (0.00) \\
\midrule
TransD
&& isPartOf (0.40) & inLanguage (0.37) && genre (0.84) & genre (0.87) \\
&& inLanguage (0.24) & isPartOf (0.31) && foundingLocation (0.13) & foundingLocation (0.11) \\
&& hasPart (0.23) & hasPart (0.23) && award (0.03) & award (0.02) \\
&& composer (0.08) & composer (0.06) && memberOf (0.00) & memberOf (0.00) \\
&& citation (0.03) & citation (0.03) && byArtist (0.00) & knowsLanguage (0.00) \\
\midrule
RESCAL
&& isPartOf (0.42) & inLanguage (0.33) && genre (0.74) & genre (0.81) \\
&& inLanguage (0.23) & isPartOf (0.33) && foundingLocation (0.22) & foundingLocation (0.16) \\
&& hasPart (0.21) & hasPart (0.25) && award (0.03) & award (0.02) \\
&& composer (0.09) & composer (0.06) && memberOf (0.01) & memberOf (0.00) \\
&& citation (0.04) & citation (0.02) && knowsLanguage (0.00) & knowsLanguage (0.00) \\
\midrule
DistMult
&& isPartOf (0.38) & inLanguage (0.39) && genre (0.81) & genre (0.86) \\
&& hasPart (0.26) & isPartOf (0.29) && foundingLocation (0.14) & foundingLocation (0.11) \\
&& inLanguage (0.22) & hasPart (0.22) && award (0.04) & award (0.02) \\
&& composer (0.08) & composer (0.05) && memberOf (0.00) & memberOf (0.00) \\
&& citation (0.05) & citation (0.03) && knowsLanguage (0.00) & knowsLanguage (0.00) \\
\midrule
TuckER
&& isPartOf (0.38) & inLanguage (0.37) && genre (0.85) & genre (0.90) \\
&& hasPart (0.25) & isPartOf (0.30) && foundingLocation (0.11) & foundingLocation (0.08) \\
&& inLanguage (0.24) & hasPart (0.23) && award (0.04) & award (0.02) \\
&& composer (0.08) & composer (0.05) && memberOf (0.01) & memberOf (0.00) \\
&& citation (0.04) & citation (0.03) && byArtist (0.00) & byArtist (0.00) \\
\midrule
BoxE
&& isPartOf (0.48) & isPartOf (0.40) && genre (0.73) & genre (0.83) \\
&& hasPart (0.29) & hasPart (0.30) && foundingLocation (0.23) & foundingLocation (0.15) \\
&& \brown{\textbf{inLanguage (0.11)}} & \brown{\textbf{inLanguage (0.19)}} && award (0.04) & award (0.02) \\
&& composer (0.10) & composer (0.07) && byArtist (0.00) & byArtist (0.00) \\
&& lyricist (0.02) & citation (0.02) && memberOf (0.00) & memberOf (0.00) \\
\midrule
ConvE
&& isPartOf (0.39) & isPartOf (0.33) && genre (0.83) & genre (0.86) \\
&& hasPart (0.25) & inLanguage (0.31) && foundingLocation (0.14) & foundingLocation (0.12) \\
&& inLanguage (0.22) & hasPart (0.25) && award (0.02) & award (0.02) \\
&& composer (0.08) & composer (0.06) && memberOf (0.01) & memberOf (0.00) \\
&& citation (0.04) & citation (0.04) && byArtist (0.00) & byArtist (0.00) \\
\bottomrule
\end{tabular}
}
\end{table}

Results for two picked classes are reported in Table~\ref{tab:expe3-results}, where each predicate is weighted in accordance to its frequency in the set of common triples between the 2-hop subgraphs of top-K neighbors in the embedding space (for $K=3$ and $K=10$) and a given entity of class \texttt{MusicComposition} and \texttt{MusicGroup} (YAGO4-19K).
A class-based analysis reveals that predicates are differently ranked, \textit{i.e.} different KGEMs \textit{may} rely on different sets of predicates. For example, for $K=10$ and entities of the class \texttt{MusicComposition}, either \texttt{isPartOf} or \texttt{inLanguage} appears as the most frequent predicate depending on the KGEM considered. 

We additionally observe that even when the ranking is the same between KGEMs -- especially since the predicate distribution can be severely skewed towards one or few predicates -- the relative importance of predicates can still differ. For instance, although \texttt{genre} is consistently ranked as the most frequent predicate for \texttt{MusicGroup} entities, results for RDF2Vec suggest a more balanced predicate distribution and a lesser emphasis put on this single predicate (highlighted in blue in Table~\ref{tab:expe3-results}). It is the only model to put a signficant emphasis on \texttt{award} (i.e., considers two \texttt{MusicGroups} as similar if they have won the same (1-hop) or similar (2-hop) awards).
In some cases, a few KGEMs seem to assign a lower relevance to specific predicates. For instance, under BoxE the top-3 and top-10 neighbors of entities that are \texttt{MusicCompositions} contain the predicate \texttt{inLanguage} at a much lesser frequency in their subgraphs' intersection (highlighted in brown in Table~\ref{tab:expe3-results}), compared to other models. Answering \textbf{RQ3}, we observe that the different notions of entity similarity reflected by different KGEMs are comparable, as most of them put a focus on the same set of predicates when determining entity similarity.

\vspace{-0.15cm}
\section{Conclusion and Outlook}\label{sec:conclusion}
\vspace{-0.05cm}

This work delved into the intricate relationship between entity similarity in KGs and their respective embeddings in KGEMs, questioning the widespread \emph{KGE entity similarity assumption}. Contrary to this belief, we showed that the choice of KGEM significantly influences the notion of entity similarity encoded in the resulting vector space. This finding has profound implications for a variety of downstream tasks where accurate entity similarity is crucial, \textit{e.g.} recommender systems and semantic searches. For instance, if the proximity of graph embeddings does not align with our proposed metric for measuring entity similarity, the tacit assumption of downstream systems exploiting embeddings for recommender systems does not hold and caution is needed when deploying such systems.

A common practice is to evaluate KGEMs by their performance in link prediction, then picking the one with the best performance for a downstream task relying on entity similarty. A critical takeaway from our study is that this practice might not yield the best results in terms of capturing true entity similarity. Instead, our results advocate for cautiousness. Moreover, in scenarios where the similarity of specific classes of entities is of paramount importance, a per-class analysis becomes essential. This approach allows for a tailored selection of KGEMs that are more adept at capturing the nuances and semantics of particular classes, thereby ensuring a more accurate and meaningful representation of entity similarity.

\bibliographystyle{splncs04}
\bibliography{references}

\begin{thebibliography}{10}
\providecommand{\url}[1]{\texttt{#1}}
\providecommand{\urlprefix}{URL }
\providecommand{\doi}[1]{https://doi.org/#1}

\bibitem{boxe}
Abboud, R., Ceylan, {\.I}.{\.I}., Lukasiewicz, T., Salvatori, T.: Boxe: {A} box
  embedding model for knowledge base completion. In: Advances in Neural
  Information Processing Systems 33: Annual Conference on Neural Information
  Processing Systems 2020, NeurIPS 2020, December 6-12, 2020, virtual (2020)

\bibitem{alshargi2018concept2vec}
Alshargi, F., Shekarpour, S., Soru, T., Sheth, A.: Concept2vec: Metrics for
  evaluating quality of embeddings for ontological concepts. arXiv preprint
  arXiv:1803.04488  (2018)

\bibitem{dbpedia}
Auer, S., Bizer, C., Kobilarov, G., Lehmann, J., Cyganiak, R., Ives, Z.G.:
  Dbpedia: {A} nucleus for a web of open data. In: The Semantic Web, 6th
  International Semantic Web Conference, 2nd Asian Semantic Web Conference,
  {ISWC} + {ASWC}. Lecture Notes in Computer Science, vol.~4825, pp. 722--735.
  Springer (2007)

\bibitem{tucker}
Balazevic, I., Allen, C., Hospedales, T.M.: Tucker: Tensor factorization for
  knowledge graph completion. In: Proceedings of the 2019 Conference on
  Empirical Methods in Natural Language Processing and the 9th International
  Joint Conference on Natural Language Processing, {EMNLP-IJCNLP} 2019, Hong
  Kong, China, November 3-7, 2019. pp. 5184--5193. Association for
  Computational Linguistics (2019). \doi{10.18653/v1/D19-1522}

\bibitem{aifb}
Bloehdorn, S., Sure, Y.: Kernel methods for mining instance data in ontologies.
  In: The Semantic Web, 6th International Semantic Web Conference, 2nd Asian
  Semantic Web Conference, {ISWC} 2007 + {ASWC} 2007, Busan, Korea, November
  11-15, 2007. Lecture Notes in Computer Science, vol.~4825, pp. 58--71.
  Springer (2007). \doi{10.1007/978-3-540-76298-0\_5}

\bibitem{transe}
Bordes, A., Usunier, N., Garc{\'{\i}}a{-}Dur{\'{a}}n, A., Weston, J.,
  Yakhnenko, O.: Translating embeddings for modeling multi-relational data. In:
  Conference on Neural Information Processing Systems (NeurIPS). pp. 2787--2795
  (2013)

\bibitem{chen2019similarity}
Chen, W., Zhu, H., Han, X., Liu, Z., Sun, M.: Quantifying similarity between
  relations with fact distribution. In: Proceedings of the 57th Conference of
  the Association for Computational Linguistics, {ACL} 2019, Florence, Italy,
  July 28- August 2, 2019, Volume 1: Long Papers. pp. 2882--2894. Association
  for Computational Linguistics (2019). \doi{10.18653/V1/P19-1278}

\bibitem{conve}
Dettmers, T., Minervini, P., Stenetorp, P., Riedel, S.: Convolutional 2d
  knowledge graph embeddings. In: Proceedings of the Thirty-Second {AAAI}
  Conference on Artificial Intelligence, (AAAI-18), the 30th innovative
  Applications of Artificial Intelligence (IAAI-18), and the 8th {AAAI}
  Symposium on Educational Advances in Artificial Intelligence (EAAI-18), New
  Orleans, Louisiana, USA, February 2-7, 2018. pp. 1811--1818. {AAAI} Press
  (2018)

\bibitem{gadelrab2020conceptual}
Gad{-}Elrab, M.H., Stepanova, D., Tran, T., Adel, H., Weikum, G.: Excut:
  Explainable embedding-based clustering over knowledge graphs. In: The
  Semantic Web - {ISWC} 2020 - 19th International Semantic Web Conference,
  Athens, Greece, November 2-6, 2020, Proceedings, Part {I}. Lecture Notes in
  Computer Science, vol. 12506, pp. 218--237. Springer (2020).
  \doi{10.1007/978-3-030-62419-4\_13}

\bibitem{guo2022survey}
Guo, Q., Zhuang, F., Qin, C., Zhu, H., Xie, X., Xiong, H., He, Q.: A survey on
  knowledge graph-based recommender systems. {IEEE} Trans. Knowl. Data Eng.
  \textbf{34}(8),  3549--3568 (2022). \doi{10.1109/TKDE.2020.3028705}

\bibitem{hubert2023sematk}
Hubert, N., Monnin, P., Brun, A., Monticolo, D.: Sem@k: Is my knowledge graph
  embedding model semantic-aware? Semantic Web  \textbf{14},  1--37 (12 2023).
  \doi{10.3233/SW-233508}

\bibitem{hubert2023maschine}
Hubert, N., Paulheim, H., Monnin, P., Brun, A., Monticolo, D.: Schema first!
  learn versatile knowledge graph embeddings by capturing semantics with
  maschine. In: {K-CAP} '23: Knowledge Capture Conference, Pensacola, Florida,
  USA, December 5-7, 2023. {ACM} (2023). \doi{10.48550/ARXIV.2306.03659}

\bibitem{ilievski2023swj}
Ilievski, F., Shenoy, K., Chalupsky, H., Klein, N., Szekely, P.: A study of
  concept similarity in wikidata. Semantic Web Journal  (2023)

\bibitem{jain2021embeddings}
Jain, N., Kalo, J.C., Balke, W.T., Krestel, R.: Do embeddings actually capture
  knowledge graph semantics? In: The Semantic Web: 18th International
  Conference, ESWC 2021, Virtual Event, June 6--10, 2021, Proceedings 18. pp.
  143--159. Springer (2021)

\bibitem{transd}
Ji, G., He, S., Xu, L., Liu, K., Zhao, J.: Knowledge graph embedding via
  dynamic mapping matrix. In: Proceedings of the 53rd Annual Meeting of the
  Association for Computational Linguistics and the 7th International Joint
  Conference on Natural Language Processing of the Asian Federation of Natural
  Language Processing, {ACL} 2015, July 26-31, 2015, Beijing, China, Volume 1:
  Long Papers. pp. 687--696. The Association for Computer Linguistics (2015).
  \doi{10.3115/v1/p15-1067}

\bibitem{ji2021survey}
Ji, S., Pan, S., Cambria, E., Marttinen, P., Yu, P.S.: A survey on knowledge
  graphs: Representation, acquisition, and applications. {IEEE} Trans. Neural
  Networks Learn. Syst.  \textbf{33}(2),  494--514 (2022).
  \doi{10.1109/TNNLS.2021.3070843}

\bibitem{kalo2019similarity}
Kalo, J., Ehler, P., Balke, W.: Knowledge graph consolidation by unifying
  synonymous relationships. In: The Semantic Web - {ISWC} 2019 - 18th
  International Semantic Web Conference, Auckland, New Zealand, October 26-30,
  2019, Proceedings, Part {I}. Lecture Notes in Computer Science, vol. 11778,
  pp. 276--292. Springer (2019). \doi{10.1007/978-3-030-30793-6\_16}

\bibitem{kendall1938}
Kendall, M.G.: A new measure of rank correlation. Biometrika  \textbf{30}(1/2),
   81--93 (1938)

\bibitem{liu2020kbert}
Liu, W., Zhou, P., Zhao, Z., Wang, Z., Ju, Q., Deng, H., Wang, P.: {K-BERT:}
  enabling language representation with knowledge graph. In: The Thirty-Fourth
  {AAAI} Conference on Artificial Intelligence, {AAAI} 2020, The Thirty-Second
  Innovative Applications of Artificial Intelligence Conference, {IAAI} 2020,
  The Tenth {AAAI} Symposium on Educational Advances in Artificial
  Intelligence, {EAAI} 2020, New York, NY, USA, February 7-12, 2020. pp.
  2901--2908. {AAAI} Press (2020). \doi{10.1609/AAAI.V34I03.5681}

\bibitem{ma2016typing}
Ma, Y., Cambria, E., Gao, S.: Label embedding for zero-shot fine-grained named
  entity typing. In: Calzolari, N., Matsumoto, Y., Prasad, R. (eds.) {COLING}
  2016, 26th International Conference on Computational Linguistics, Proceedings
  of the Conference: Technical Papers, December 11-16, 2016, Osaka, Japan. pp.
  171--180. {ACL} (2016), \url{https://aclanthology.org/C16-1017/}

\bibitem{mikolov2013}
Mikolov, T., Chen, K., Corrado, G., Dean, J.: Efficient estimation of word
  representations in vector space. In: 1st International Conference on Learning
  Representations, {ICLR} 2013, Scottsdale, Arizona, USA, May 2-4, 2013,
  Workshop Track Proceedings (2013)

\bibitem{rescal}
Nickel, M., Tresp, V., Kriegel, H.: A three-way model for collective learning
  on multi-relational data. In: Proc. of the 28th International Conference on
  Machine Learning, {ICML}. pp. 809--816 (2011)

\bibitem{portisch2022knowledge}
Portisch, J., Heist, N., Paulheim, H.: Knowledge graph embedding for data
  mining vs. knowledge graph embedding for link prediction--two sides of the
  same coin? Semantic Web  \textbf{13}(3),  399--422 (2022)

\bibitem{ristoski2016rdf2vec}
Ristoski, P., Paulheim, H.: Rdf2vec: {RDF} graph embeddings for data mining.
  In: The Semantic Web - {ISWC} 2016 - 15th International Semantic Web
  Conference, Kobe, Japan, October 17-21, 2016, Proceedings, Part {I}. Lecture
  Notes in Computer Science, vol.~9981, pp. 498--514 (2016).
  \doi{10.1007/978-3-319-46523-4\_30}

\bibitem{rossi2021survey}
Rossi, A., Barbosa, D., Firmani, D., Matinata, A., Merialdo, P.: Knowledge
  graph embedding for link prediction: {A} comparative analysis. {ACM}
  Transactions on Knowledge Discovery from Data  \textbf{15}(2),  14:1--14:49
  (2021)

\bibitem{rossi2020relations}
Rossi, A., Matinata, A.: Knowledge graph embeddings: Are relation-learning
  models learning relations? In: Proceedings of the Workshops of the
  {EDBT/ICDT} 2020 Joint Conference, Copenhagen, Denmark, March 30, 2020.
  {CEUR} Workshop Proceedings, vol.~2578. CEUR-WS.org (2020)

\bibitem{codex}
Safavi, T., Koutra, D.: Codex: {A} comprehensive knowledge graph completion
  benchmark. In: Proceedings of the 2020 Conference on Empirical Methods in
  Natural Language Processing, {EMNLP} 2020, Online, November 16-20, 2020. pp.
  8328--8350. Association for Computational Linguistics (2020).
  \doi{10.18653/v1/2020.emnlp-main.669}

\bibitem{sanfeliu1983}
Sanfeliu, A., Fu, K.: A distance measure between attributed relational graphs
  for pattern recognition. {IEEE} Trans. Syst. Man Cybern.  \textbf{13}(3),
  353--362 (1983). \doi{10.1109/TSMC.1983.6313167}

\bibitem{shi2018conmask}
Shi, B., Weninger, T.: Open-world knowledge graph completion. In: Proceedings
  of the Thirty-Second {AAAI} Conference on Artificial Intelligence, (AAAI-18),
  the 30th innovative Applications of Artificial Intelligence (IAAI-18), and
  the 8th {AAAI} Symposium on Educational Advances in Artificial Intelligence
  (EAAI-18), New Orleans, Louisiana, USA, February 2-7, 2018. pp. 1957--1964.
  {AAAI} Press (2018). \doi{10.1609/AAAI.V32I1.11535}

\bibitem{sosa2020}
Sosa, D.N., Derry, A., Guo, M.G., Wei, E., Brinton, C., Altman, R.B.: A
  literature-based knowledge graph embedding method for identifying drug
  repurposing opportunities in rare diseases. In: Pacific Symposium on
  Biocomputing 2020, Fairmont Orchid, Hawaii, USA, January 3-7, 2020. pp.
  463--474 (2020)

\bibitem{yago}
Suchanek, F.M., Kasneci, G., Weikum, G.: Yago: a core of semantic knowledge.
  In: Proc. of the 16th International Conference on World Wide Web, {WWW}. pp.
  697--706. {ACM} (2007)

\bibitem{sun2020ernie}
Sun, Y., Wang, S., Li, Y., Feng, S., Tian, H., Wu, H., Wang, H.: {ERNIE} 2.0:
  {A} continual pre-training framework for language understanding. In: The
  Thirty-Fourth {AAAI} Conference on Artificial Intelligence, {AAAI} 2020, The
  Thirty-Second Innovative Applications of Artificial Intelligence Conference,
  {IAAI} 2020, The Tenth {AAAI} Symposium on Educational Advances in Artificial
  Intelligence, {EAAI} 2020, New York, NY, USA, February 7-12, 2020. pp.
  8968--8975. {AAAI} Press (2020). \doi{10.1609/AAAI.V34I05.6428}

\bibitem{sun2020entity}
Sun, Z., Zhang, Q., Hu, W., Wang, C., Chen, M., Akrami, F., Li, C.: A
  benchmarking study of embedding-based entity alignment for knowledge graphs.
  Proc. {VLDB} Endow.  \textbf{13}(11),  2326--2340 (2020)

\bibitem{sun2020survey}
Sun, Z., Zhang, Q., Hu, W., Wang, C., Chen, M., Akrami, F., Li, C.: A
  benchmarking study of embedding-based entity alignment for knowledge graphs.
  Proc. {VLDB} Endow.  \textbf{13}(11),  2326--2340 (2020)

\bibitem{rotate}
Sun, Z., Deng, Z., Nie, J., Tang, J.: Rotate: Knowledge graph embedding by
  relational rotation in complex space. In: 7th International Conference on
  Learning Representations, {ICLR} (2019)

\bibitem{toutanova}
Toutanova, K., Chen, D.: Observed versus latent features for knowledge base and
  text inference. In: Proc. of the 3rd Workshop on Continuous Vector Space
  Models and their Compositionality. pp. 57--66. Association for Computational
  Linguistics (2015)

\bibitem{complex}
Trouillon, T., Welbl, J., Riedel, S., Gaussier, {\'{E}}., Bouchard, G.: Complex
  embeddings for simple link prediction. In: Proceedings of the 33rd
  International Conference on Machine Learning, {ICML}. vol.~48, pp. 2071--2080
  (2016)

\bibitem{wang2021survey}
Wang, M., Qiu, L., Wang, X.: A survey on knowledge graph embeddings for link
  prediction. Symmetry  \textbf{13}(3), ~485 (2021)

\bibitem{webber2010rbo}
Webber, W., Moffat, A., Zobel, J.: A similarity measure for indefinite
  rankings. {ACM} Trans. Inf. Syst.  \textbf{28}(4),  20:1--20:38 (2010).
  \doi{10.1145/1852102.1852106}

\bibitem{distmult}
Yang, B., Yih, W., He, X., Gao, J., Deng, L.: Embedding entities and relations
  for learning and inference in knowledge bases. In: 3rd International
  Conference on Learning Representations, {ICLR} (2015)

\end{thebibliography}

\end{document}